\documentclass[11pt, a4paper, singlecolumn, logo]{deepmind}
\pdfoutput=1

\pdftrailerid{redacted}

\makeatletter
\renewcommand\bibentry[1]{\nocite{#1}{\frenchspacing\@nameuse{BR@r@#1\@extra@b@citeb}}}
\makeatother

\usepackage{kantlipsum, lipsum}
\usepackage{dsfont}
\usepackage{dm-colors}
\usepackage{caption}
\usepackage{newfloat}

\DeclareCaptionType{Box}

\usepackage[numbers]{natbib} 

\graphicspath{{figures/}}

\title{Symbolic Behaviour in Artificial Intelligence}

\correspondingauthor{adamsantoro@google.com}

\paperurl{}

\reportnumber{} 

\renewcommand{\today}{} 

\author[*,1]{Adam Santoro}
\author[*,1]{Andrew Lampinen}
\author[1]{Kory W. Mathewson}
\author[1]{Timothy Lillicrap}
\author[1]{David Raposo}

\affil[*]{Equal contributions}
\affil[1]{DeepMind}

\begin{abstract}
The human ability to use symbols has yet to be replicated in machines. Bridging the gap requires considering how symbol meaning is established: if it is symbol users who agree-upon symbol meaning, then symbol-use comprises behaviours that navigate agreements about meaning. We leverage this insight to articulate graded symbolic behaviours, including constructing new symbols, altering prior ones, and introspecting about meaning and reasoning processes. We then evaluate contemporary AI methods along each criterion. Ultimately, we argue that fluent symbol-use in machines will emerge when learning-based agents immerse in human socio-cultural interactions demanding coordination around perspectives and meaning. By collecting these scenarios at scale, and including interactive human feedback, researchers can optimize towards symbolic fluency using contemporary algorithms for moulding behaviour.
\end{abstract}

\begin{document}

\maketitle

\section{Symbols and Intelligence}
The ability to proficiently use symbols makes it possible to reason about the laws of nature, fly to the moon, write poetry, and evoke thoughts and feelings in the minds of others. Artificial intelligence (AI) pioneers Newell and Simon claimed that ``[s]ymbols lie at the root of intelligent action'' and should therefore be a central component in the design of artificial intelligence~\citep{newell2007computer}. Their hypothesis drove a decades-long program of Good Old-Fashioned AI (GOFAI) research, which attempted to create intelligent machines by applying the syntactic mechanisms developed in computer science, logic, mathematics, linguistics, and psychology.

Much contemporary artificial intelligence (AI) research strays from GOFAI methods and instead leverages learning-based artificial neural networks~\citep{lecun2015deep}. Neural networks have achieved substantial recent success in domains like language \citep{brown2020language,wei2021finetuned} and mathematics \citep{lample2019deep}, which were historically thought to require classical symbolic approaches. Despite these successes, some apparent weaknesses of current connectionist-based approaches \citep{lake2018generalization,kemker2018measuring,zhou2020evaluating} have lead to calls for a return of classical symbolic methods, possibly in the form of hybrid, neuro-symbolic models~\citep{marcus2018deep,marcus2020next,lake2017building,garcez2020neurosymbolic,garnelo2019reconciling}.  

We suggest a different way forward that leverages a perspective on symbols originating in philosophy and semiotics~\citep{hartshorne1931collected}, rather than classical logic and computer science~\citep[cf.][]{taniguchi2018symbol}. This perspective emphasizes the \emph{triadic relation of meaning}: Symbols are (1) \emph{entities} that (2) mean \emph{something} (3) to \emph{someone}~\citep{de2007peirce}. An interpreter (the aforementioned \emph{someone}) exhibits an ability to come to an agreement---with interlocutors, or simply internally---about the arbitrary link between a substrate and what it denotes. 

The subjectivity of symbol meaning motivates viewing symbolic fluency through a distinctly behavioural lens. This perspective is reminiscent of Wittgenstein's proposal that we ought to understand language by examining how words are used; i.e., the associated behaviours (\citealp{wittgenstein1958blue}; cf. \citealp{lenci2018distributional,boleda2020distributional}). Symbol use constitutes a subclass of behaviours that leverage the knowledge that symbol meaning is purely \emph{conventional}~\citep{lewis1969convention}, and that symbol meaning can, and should, be invented and continuously adapted.

Consequently, instead of focusing dichotomously on \emph{whether} a system (human, animal, or AI) engages with symbols, we focus on characterizing \emph{how} it engages with symbols; that is, \emph{how} it exhibits behaviours that implicate  meaning-by-convention. This focus offers a set of behavioural criteria that outline the varieties and gradations of symbolic behaviour. These criteria are measurable, which is useful both for assessing current AI, and as a direct target for optimization.

The origin and development of symbolic behaviour in humans suggests a way to make progress towards developing AI that engages with symbols as humans do. Just as human symbolic fluency is built upon an ability to coordinate and communicate in social settings~ \citep{deacon1998symbolic,tomasello2019becoming}, we propose subjecting learning-based machines to similar socio-cultural pressures, for example by learning from large datasets of human behaviours and by leveraging online interaction with humans and other agents~\citep{abramson2020imitating}. Ultimately, it is through the combination of rich, challenging, diverse experiences of human-like socio-cultural interactions and powerful learning-based algorithms that we will develop machines that proficiently use symbols.

\section{Symbols Are Subjective}
\label{sec:properties}
Newell and Simon define symbols as a set of interrelated ``physical patterns'' that could ``designate any expression whatsoever''~\citep{newell2007computer}. But this definition---and how it is used in practice---is controversial. For example, \citet{touretzky1994reconstructing} expressed frustration that some researchers render the concept of symbol operationally vacuous by claiming that \emph{anything} that designates or denotes is a symbol.

The philosopher Charles Sanders Peirce, many decades earlier, provided a complementary view. Peirce first outlined three categories of relation to establish meaning: icons, indices, and symbols~\citep{deacon1998symbolic,hartshorne1931collected}. Unlike icons (which make reference by way of similarity) or indices (which leverage some temporal or physical connection to the things to which they refer), symbols depend solely upon an ``agreed upon link'' to establish meaning. Thus, meaning does not depend on the actual physical or temporal characteristics of the medium. It is only by agreement that a flag comes to symbolize one country rather than another, and there is nothing inherent in the properties of cotton or ink that reveals the distinction. 

Newell, Simon, and Peirce therefore agree that a symbol consists of some substrate---a scratch on a page, an air vibration, or an electrical pulse---that is imbued with arbitrary meaning, but Peirce emphasizes the \emph{subjectivity} of this convention. A substrate is only a symbol with respect to some interpreter(s), making symbols an irreducibly triadic phenomenon whose meanings are neither intrinsic to their substrate, nor objective~\citep{de2007peirce}. 

Because symbols are inherently subjective constructs, their use and meaning is entangled with the state of their interpreters. As an interpreter's background, capabilities, and biases vary, so too will their subjective treatment of symbols. This makes it problematic to draw binary distinctions between using and not using symbols. Indeed, as we will explore, symbol use manifests differently over the course of development, as broader cognitive systems and knowledge structures change \citep[cf.][]{mcclelland2010letting}. In the next section, we therefore outline graded criteria for assessing various aspects of \emph{symbolic behaviour}---engagement with meaning-by-convention.

\section{Symbolic Behaviour}
\label{sec:symbolic_thinking}

An interpreter's \emph{symbolic behaviour}---their use of symbols---roughly comprises a set of interrelated traits that reveal how they participate in an infrastructure of meaning-by-convention. In particular, human symbolic behaviour is \emph{receptive, constructive, embedded, malleable, meaningful, and graded}. In this section we elaborate on each dimension, and evaluate the progress of contemporary AI.

\textbf{Receptive}. Symbolic behaviour includes the ability to appreciate existing conventions, and to receive new ones. For example, humans can learn a new word from a definition or example. But many animals and models can learn such associations to some degree.
So, while being receptive is necessary for participation in a symbolic framework, it is not sufficient for human-level symbolic fluency.

Receptiveness is the predominant trait of current AI. For example, language models learn to imitate existing conventions \citep{brown2020language}, and multi-modal models learn about conventions of denotation, so that they can e.g. produce an image from a description~\citep{radford2learning}. Instruction-following agents map between symbolic inputs and goals~\citep{hermann2017grounded}. Proponents of neuro-symbolic models often emphasize these models' ability to rapidly learn a new concept, from a definition or a few examples \citep[e.g.][]{lake2015human, marcus2020next,zhou2021flexible}. Deep learning models can also be receptive to established convention with only a single exposure (or a few exposures) to a new symbol definition~\citep{hill2020grounded,brown2020language}. When transformers are trained to construct images from captions, these models can receive a composition of known concepts and generate plausible renderings of novel ones, such as a ``daikon radish in a tutu walking a dog'' \citep{radford2learning}. Evidently, the ability to engage with established convention is broadly represented in AI research. 

\textbf{Constructive}. A second trait of symbolic behaviour is the ability to form new conventions; \emph{because} meaning is conventional, it can be imposed arbitrarily on top of any substrate. Such an ability can be used to increase the efficiency of communication or reasoning (e.g. by creating a new term for a recurring situation) and for creating new systems of knowledge. Consider, for example, the scientific understanding which followed the introduction of the concept of the ``genes'' to describe units of heritable variability.

Evidence for this behavioural trait is less common in current AI research. Few-shot learning~\citep{wang2020generalizing} explores a model's ability to rapidly assign meaning to a new substrate, but the meaning is pre-determined by humans. Some neuro-symbolic or classical models can construct abstractions that capture some useful structure \citep[e.g.][]{ellis2020dreamcoder,piantadosi2020computational}, but so far these abilities have only been demonstrated within very simplified and algebraically-structured domains. Deep learning models can also create abstractions, e.g. summaries of documents \citep{khandelwal2019sample} or code \citep{ahmad2020transformer}. Some ability to construct novel concepts emerges in large language models (see Box \ref{sec:language_models_box}), but only when prompted. Stronger evidence for this behavioural trait might look like the following: (i) A model is asked a question, and when explaining its answer it invents a new symbol to clarify its reasoning process or communication. (ii) Agents cooperating on a team invent terminology so that they can communicate without exposing their strategy to their opponents. (iii) A model constructs a novel mathematical concept in order to prove something new (cf. Box \ref{sec:syntax_mathematics_box}.)

\textbf{Embedded}. Symbols conform and contribute to the broader symbolic system in which they’re situated, and symbolic behaviour reflects this understanding \citep{deacon1998symbolic}. Because the meaning of a symbol is determined partly through its interactions with the entire symbol system,introducing new symbols can radically alter the interpretation of other symbols. For example, the discovery of category theory as a unifying perspective on many mathematical research areas fundamentally changed the questions researchers ask~\citep{mac1988concepts}. Humans use embodied understanding, e.g. gestures, to help them grasp abstract concepts \citep{goldin1999role,congdon2017better}, and domain knowledge aids logical reasoning in Wason Selection Task analogues that frame a logical problem in terms of a social situation \citep[][]{johnson1972reasoning,sperber1995relevance,fiddick2000no}. 
The whole structure of knowledge in which symbols are embedded can, and usually does, affect symbol-use. 

Current AI models demonstrate some embedded behaviour. Neural networks in particular have strong representational and functional biases toward such behaviour. For example, continuous vector representations are meaningful with respect to the magnitudes and angles of other vectors, and such vector relations are directly shaped by gradient-based learning to be useful for a downstream task, such as classification~\citep{devlin2019bert}. For this reason, deep learning models can suffer from poor generalization~\citep{lake2018generalization} without sufficient inductive biases, or data from which to learn the interrelationships between symbols. However, performance generally improves when symbols are embedded within a richer, more embodied contexts \citep{hill2019environmental,mcclelland2020placing}, or even using auxiliary data to learn symbol-symbol relations, as is accomplished with pre-trained word embeddings~\citep{furrer2020compositional}. These continuous vectors capture complex relationships like analogies \citep{mikolov2013efficient}, providing concrete evidence for the utility of this embedded property (though there are also drawbacks, such as capturing biases in language use~\citep{bolukbasi2016man}).

\textbf{Malleable}. Fluent symbol users understand that the conventionality of meaning allows for change. Meaning can be altered by context, by the creation of other symbols (see ``embedded'', above) or concepts, or by intentionally redefining the symbol.

The collaborative and pragmatic nature of language exemplifies malleability~\citep{clark1986referring,hawkins2017convention}. Language interpretation depends heavily on the shared situation and the mutual understanding between the speaker and listener, and so understanding human communication requires an ability to continually form pragmatic conventions~\citep{clark1986referring,hawkins2018emerging}. Beyond a single situation, malleability is often used in constructing new symbols too as language evolves---for example, in the way that ``computer mouse'' constructs a new concept by building on and transforming the existing word ``mouse.''

Furthermore, permanently altering the meaning of an existing symbol can be useful. Deep insight can require epistemic humility---recognizing that symbol meaning \emph{could} be otherwise, or \emph{should} be otherwise. For example, extending the concept of numbers to include complex numbers allowed humans to mathematically describe new phenomena. To reason flexibly, it is necessary to reshape extant meanings that are misaligned with the world or other symbols. But to do this, one must first appreciate that meaning \emph{can} be reshaped because it is established by convention. 

It's not obvious whether any AI system exhibits malleable understanding of symbols as humans do. Language models can discriminate different word senses \citep{loureiro2021analysis} and exhibit some aspects of pragmatic understanding \citep{jiang-de-marneffe-2019-evaluating}. These models demonstrate impressive abilities to learn the many subtle constraints that determine language meaning in context, and will likely improve when they are augmented with more human-like faculties and grounded experience \citep{mcclelland2020placing}. However, it's less clear whether these models exhibit behaviour that demonstrates that they can, or should, decide to actively shift their understanding of an already known symbol. Any learning-based models will surely alter its understanding of a symbol as it experiences new data. However, this type of malleability is passive, as it depends on researchers to provide the data from which new meaning could be derived. Humans, on the contrary, are malleable \emph{with purpose}, whether that purpose is to permit more fluid communication, or to come to a deeper understanding of some phenomenon. Such intentional action motivates techniques that exploit the situated, goal-directed aspects cognition, such as reinforcement learning~\citep{sutton1998introduction,stiennon2020learning,silver2021reward}.

\textbf{Meaningful}. Symbolic behavior should demonstrate use and communication of the meaning behind the symbols \emph{and} the reasoning processes that employ them. While \citet{newell1980physical} advocated for purely syntactic manipulations, we emphasize that syntactic operations do not consider the steps where meaning is extracted from, and later imbued in, arbitrary physical tokens. Focusing on syntactic manipulations assumes that such manipulations would be useful to a system that has yet to demonstrate an understanding of symbols \emph{as symbols}. Instead, even in logical domains like mathematics, meaning dominates human reasoning (Box \ref{sec:syntax_mathematics_box}). Analogously, \citet{marder2020living} describes theory in biology as ``disciplined dreaming'' wherein we face the ``challenge of creatively marrying the rules of mathematics and physics with what is known of fundamental biological principles.''    

However, meaning is more than a heuristic for syntactic reasoning. Models must understand their reasoning processes themselves as meaningful. \citet{harnad1990symbol} alluded to this fact in his focus on grounding tokens, noting that ``the syntactic manipulations both actual and possible and the rules [must be] `semantically interpretable.''' The principles by which the model manipulates its symbols must themselves be understood symbolically by the system.

Meaningful symbolic behaviour therefore involves both using semantics to reason more effectively (e.g. finding a proof of a theorem efficiently rather than blindly searching) and communicating how the semantics licenses the reasoning process (e.g. why a particular proof technique is useful, or which steps are critical). Many recent models have used deep learning to partially address the first point. For example, AlphaZero \citep{silver2017mastering} uses learned move value estimates as strong heuristics for reducing its search space. However, because AlphaZero's reasoning process (Monte-Carlo Tree Search) is hand engineered and strictly rule-based, it cannot achieve the second goal. Analogous limitations apply to neuro-symbolic models that use deep learning within a more programmatic reasoning process \citep[e.g.][]{ellis2020dreamcoder, nye2021improving}. This motivates future research toward systems that can understand their reasoning processes as meaningful, and use that understanding to refine and communicate their reasoning.

Meaningful reasoning is central to critical issues of safety and ethics. For example, we echo others who highlighted the challenges of creating rule-based machine ethics \citep[e.g.][]{allen2005artificial,brundage2014limitations}. Without understanding the meaning behind the rules, such systems would only follow the letter of the law, not the spirit. Consider a rule like ``don't discriminate on the basis of race.'' As US history unfortunately illustrates, it's easy to find a proxy variable for race, like neighborhood, and have essentially the same discriminatory effect \citep{woods2012federal}. 
We need a system that behaves in accordance with the meaning behind its principles---a system with judgement~\citep{smith2019promise}. We suggest that this goal will require pursuing AI that can exhibit symbolic behavior within a holistic, meaningful framework of ethics.

\textbf{Graded}. Finally, symbol use is not binary \citep[c.f.][]{mcclelland2010letting,mcclelland2016parallel}. Each of the behavioural criteria is best described as a spectrum of competencies. For example, young children might be receptive to simple language conventions, but unable to construct useful conventions themselves. Similarly, few humans may be able to identify certain useful alterations to meaning, because most may lack the relevant knowledge and experience (consider how scientific insights, such as a heliocentric solar system, are often initially conceived by only one or a few individuals).

Symbolic thinking is therefore not an independent, isolated function, but is rather an interdependent component of our broader cognition. And since other aspects of our cognition---attention, memory, perception---can be graded, so too can our ability to think symbolically. As it regards AI, acknowledging that the ability to use symbols is graded entails acknowledging that symbol use is not synonymous with the existence a particular computation (e.g. syntactic control over entities), or the use of a particular type of representation (e.g. discrete tokens), see Box \ref{sec:reconciling_box}. We argue instead for characterizing behaviourally \emph{how} a system expresses engagement with symbols. If certain computations or representations are essential, this should be demonstrated through the behavioural competence of a system that includes them when performing rich tasks. Thus, in the next section we draw inspiration from the development of symbolic behaviour in humans to suggest a path towards achieving more human-like symbolic behaviour in artificial intelligence.

\noindent\fbox{
\parbox{\textwidth}{
\captionof{Box}{\textbf{Symbolic behaviour in large language models}}\label{sec:language_models_box} \vspace{-0.6em}\hrule\vspace{0.6em}

Language was traditionally regarded as a quintessentially GOFAI domain. However, neural networks without symbolic components have advanced rapidly in natural language  (NLP). Transformers \citep{vaswani2017attention}---models that use attention to process a sequence---are achieving particularly remarkable successes \citep[e.g.][]{raffel2020exploring,brown2020language}. When trained to simply predict the next word (or word-piece) on internet text, these models demonstrate flexibility: they perform passably on tasks like translation, difficult Winograd schemas \citep{sakaguchi2020winogrande}, and programming problems \citep{austin2021program}. They do so even \emph{without explicit training on those tasks}, from a task description or a few examples \citep{raffel2020exploring,brown2020language}. When fully trained on translation \citep{stahlberg2020neural}, programming \citep{austin2021program}, or mathematics (see \ref{sec:syntax_mathematics_box}), transformers substantially outperform existing symbolic or hybrid models. Furthermore, the representations in transformer models are highly predictive of human brain activity during language processing \citep{schrimpf2020neural}, suggesting that they capture some aspects of human language processing. These models also achieve some symbolic behaviour, which we assess below.

\textbf{Receptive:} Large models can correctly conjugate and use a novel verb after receiving a definition \citep{brown2020language}. 

\textbf{Malleable:} Language models exhibit some pragmatics \citep[e.g.][]{loureiro2021analysis}. However, they do not exhibit intentionally malleable behaviour, such as changing meanings to propose a better definition.

\textbf{Constructive:}
We contributed a test of constructing a concept uniting disparate concepts (e.g. theoretical particle physics, tennis rackets, and sewing all involve strings) to the BIG-Benchmark \citep[][``novel\_concepts'' task]{brain2021big}, and language models exhibit some success. But, there is little evidence that these models could devise useful constructions in the pursuit of some goal.

\textbf{Embedded \& meaningful:} Despite arguments that models cannot understand meaning from language alone \citep{bender2020climbing}, recent research shows that transformers do build implicit models of situations, entities, and relations \citep{li2021implicit}. Thus, it is possible to learn aspects of meaning from language alone \citep[cf.][]{lenci2018distributional,boleda2020distributional}, though these models would still benefit from embedding in more human-like capacities and grounding (\citealp{mcclelland2020placing}; cf. \citealp{lakoff2008metaphors}).

However, there is little evidence that language models understand their own reasoning processes as meaningful or intentional. Transformers can sometimes manipulate their own outputs, e.g. reducing bias after reprocessing \citep{schick2021self} or correcting a program they produced with a human explanation of their mistakes \citep{austin2021program}. But, to exhibit these behaviours, the system designers need to carefully engineer the way the outputs are processed and reprocessed by the system; the model lacks the understanding and control necessary to manipulate its own reasoning.

\textbf{Graded:} In summary, these models exhibit some gradations of symbolic behaviour, but not full symbolic fluency. These models lack human-like embedding, and are mostly passive recipients, rather than constructing or changing symbols in pursuit of some goal. These limitations would be addressed by the interactive, socio-cultural immersion we suggest below.
}}

\noindent\fbox{
\parbox{\textwidth}{
\captionof{Box}{\textbf{Syntax \& meaning in mathematics}} \label{sec:syntax_mathematics_box}\vspace{-0.6em}\hrule\vspace{0.6em}
Newell \& Simon motivated the physical symbol system hypothesis by this claim: ``Logic, and by incorporation all of mathematics, was a game played with meaningless tokens according to certain purely syntactic rules. All meaning had been purged.'' \citep{newell2007computer}. In the reasoning systems they created, symbols thus became synonymous with discrete entities, symbolic processing entailed manipulating with formal algebras, and the properties of our formal symbolic systems were inherited by GOFAI. But, is this how humans perform logical reasoning? Fields medalist Paul Cohen responded directly: ``I never was able to successfully analyze proofs as a combinatorial `game' played with symbols on paper,'' instead, to reason productively ``one must essentially forget that all proofs are eventually transcribed in this formal language'' \citep{cohen2002discovery}.  \citet{maclane2012mathematics} explains why: ``strict formalism can't explain which of many formulas matter [...] the choice of form is determined by ideas and experience.'' The meaning of symbols and their relationships gives mathematicians intuitions for why a theorem might be true, and thereby ideas of how to prove it (as do examples; \citealp{polya1990mathematics}). Furthermore, a major goal of theories and proofs is to convey understanding. Thus, mathematicians are often unsatisfied with a formal proof of a theorem that does not clearly communicate understanding of the underlying ideas \citep{mac1988concepts,lange2015depth}. The meaning behind symbols is the essence of mathematics, not a burden to be discarded. 

\citet{mcclelland2016parallel} draw on similar arguments to suggest that neural network models are uniquely well-posed to capture the meaning-laden processes of human mathematical reasoning. Indeed, deep learning models are increasingly succeeding at difficult mathematical reasoning tasks like identifying the theorem needed to complete a proof \citep{li2020isarstep}, or solving difficult symbolic integration \citep{lample2019deep}, without GOFAI or neuro-symbolic methods. \citet{lample2019deep} even show that transformers in some cases outperform classical symbol systems like Mathematica that were specifically designed for these tasks, though there are certainly caveats to that particular comparison \citep{davis2019use}. Deep learning models will likely improve further when they can learn from a more human-like curriculum---even simple reasoning curricula can improve these models performance \citep{wu2021lime}. Similarly, while it is difficult for transformers to learn to perform division from problems and answers alone, even with millions of examples \citep{saxton2018analysing}, a pretrained language model endowed with a virtual notepad and trained on \emph{only 200} detailed demonstrations of long division can achieve 80\% generalization to new problems \citep{recchia2021teaching}. Even beyond the curriculum, humans build from physical experience to abstractions \citep{lakoff2000mathematics, lakoff2008metaphors}, e.g. gestures help us learn mathematics \citep{congdon2017better,goldin1999role}, and deep learning might likewise benefit from such grounding.

Furthermore, current models do not generally reason about their own reasoning processes, or interpret them as \emph{meaningful}. We suggest that achieving this will be important for achieving fully human-like mathematical abilities.}}

\section{How Can Symbolic Behaviour Come About?}

Some traditional theories suggest that symbolic competency could be enabled by unique events in evolution that equip humans with ``innate'' mental organs.~\citep{yang2017growth,fodor1975language}. However, other theories posit that the human ability for symbolic thought may have instead resulted from a gradual internalization of evolved behaviours involved in socio-cultural interaction and communication~\citep{de2007peirce,deacon1998symbolic,lazaridou2020emergent,kirby2002learning}. That is, the cognitive requirements for outwardly establishing and appreciating meaning-by-convention among a set of interlocutors consequently permit internal symbolic thought, setting up a virtuous, incremental cycle of co-development~\citep{deacon1998symbolic,de2007peirce}.

This process may have occurred on an evolutionary time-scale in humans, but even within single lifetimes humans gain critical experience learning the consequences of meaning-by-convention. For example, children must learn that words will be known by speakers of the same language (\emph{but not others}), and if they are bilingual, they also learn how to not mix languages~\citep{wei2020bilingualism}. They also learn to consider speaker knowledge and intent~\citep{diesendruck2005principles}, reinterpret or ignore noun-object bindings if a speaker is unreliable~\citep{dautriche_goupil_smith_rabagliati_2020}, and infer social norms from others' behaviour~\citep{schmidt2016young}.  \citet{tomasello2019becoming} argues persuasively that the ability to reconcile different perspectives is prerequisite to the ability to manage convention, and hence, to coordinate on symbol meaning and use. 

Analogously, fluent symbol-use in machines will emerge when situated, learning-based agents are immersed in scenarios that demand an active participation with meaning-by-convention. Compared to pure multi-agent settings, immersion in human socio-cultural situations forces agents to cope with, and learn from, the diverse symbolic behaviours that humans already exhibit. Practically, human experiences can be compiled into enormous datasets from which agents can imitate rich symbolic behaviours, and human-agent interactions can be gathered in real-time so that humans can provide behavioural feedback signals that cannot be programmed \emph{a priori}. In both cases, these experiences can be obtained at scale, which allows for the use of supervised, self-supervised, and reinforcement learning to learn human-like symbolic behaviours.

Some recent work has begun to pursue directly optimizing natural human-agent interactions at scale. In \citep{abramson2020imitating}, humans and agents take control of avatars in a virtual 3D environment called a ``Playroom'', comprising objects such as furniture and toys. Setter agents (or humans) pose a task for Solver agents (or humans) using natural language, such as ``pick up the blue duck and bring it to the bedroom''.  Using imitation learning techniques, the authors were able to train artificial agents to display impressive aspects of symbolic behaviour, such as embeddedness (consider how the meaning of ``duck'' co-depends on other symbols, perceptions, motor behaviours, and goals) and malleability (to handle coreferents like ``it'' and ``that''). We see this as a promising approach to placing agents in situations requiring conventional, perspectival, and pragmatic interaction with humans. Critically, the authors collected a vast quantity of data---the agents learned from more than 600,000 episodes of human-human interaction in a complex environment. Collecting rich compilations of symbolic behaviours at scale allows for direct optimization towards these competencies. However, further work is needed to create experiences that develop particular aspects of symbolic behaviour, such as constructive and meaningful behaviours.

This argument aligns with a broad perspective in contemporary 
AI that robust cognitive faculties emerge because we optimize learning-based systems for desired behaviours directly, rather than designing internal mechanisms that may ultimately prove limiting \citep{hansen2017building,silver2021reward}. It has roots in cognitive science~\citep[e.g.][]{mcclelland2010letting,mcclelland1986parallel}, and is now more broadly supported by findings of recent research in AI: researchers often progressively remove designed knowledge in favor of the power of general learning systems, for example from AlphaGo~\citep{silver2016mastering}, to AlphaZero \citep{silver2017mastering}, and MuZero~\citep{schrittwieser2020mastering}. Moreover, even if some knowledge or ability is ``innate'' in humans, it must have emerged through evolutionary pressures acting on experiences, which can be captured by meta-learning or other learning-based algorithms in artificial settings~\citep{wang2021meta}. Indeed, evolution-like competition and selection of models across a few generations has supported many recent succeses of RL \citep[e.g.][]{vinyals2019grandmaster,team2021open} and language \citep[e.g.][]{so2021primer}.

The idea that aspects of intelligence are emergent is not new \citep[e.g.][]{anderson1972more,mcclelland2010emergence}, but recent findings in AI reinforce that scale (both of models and data) may be critical to achieving human-like behaviour---the lesson that ``more is different''  \citep{anderson1972more} is true in artificial intelligence as well. 
For example, larger models learn more than smaller models from the same amount of data \citep{kaplan2020scaling}, and in some cases exhibit qualitatively different behaviors, such as acquiring new capacities for reasoning \citep[][Fig. 3.10]{brown2020language} or generalization \citep[][Fig. 6B]{wei2021finetuned}. Scale is not the only factor, however; interactive feedback can be more valuable than orders of magnitude of scale \citep{stiennon2020learning}. Furthermore, generalization can be qualitatively different when experience is more interactive or embodied \citep{hill2019environmental}, and training tasks are more diverse \citep{team2021open}. 
Immersion in human culture comprises the best of all these worlds because it can dramatically increase the scale, richness, diversity, and interactivity of our agents' experiences.

\noindent\fbox{
\parbox{\textwidth}{
\captionof{Box}{\textbf{Reconciling our view with other perspectives}}\label{sec:reconciling_box}\vspace{-0.6em}\hrule\vspace{0.6em}
The ongoing debate about symbols in AI mostly revolves around the properties of certain ways of computing~\citep{marcus2018deep,marcus2020next,lake2017building,garcez2020neurosymbolic,garnelo2019reconciling}. By contrast, our behavioural perspective does not prescribe the functions or mechanisms required for symbolic fluency. Advocates for GOFAI-like mechanisms often also cite behavioural evidence, particularly humans' relatively systematic compositional generalization, which is often assumed to rely on GOFAI-like compositional mechanisms \citep[e.g.][]{fodor1988connectionism}. Our views differ in several crucial ways (see also \citealp{potts2019case,hill2019environmental}).\\

First, under our interpretation symbol-use does not entail perfect systematicity. We acknowledge the graded, flawed nature of human behaviour---even adult humans are often far from perfectly systematic, achieving compositional generalization performance of only 70-90\% even within relatively simple domains \citep{lake2019human}. Furthermore, we emphasize that systematic behaviour does not comprise a more pure mode of engaging with symbols. In some cases, such approaches could even be detrimental; natural language, for example, is probably not strictly compositional~\citep{fodor2001language,sep-compositionality}. Enforcing strict algebraic notions of compositionality \emph{a priori} will make it impossible to capture the full structure of such domains. Indeed, the graded compositionality afforded by neural networks may play a role in their superior performance on language understanding compared to systems with stricter assumptions \citep{mcclelland2020placing}. Our perspective embraces these lessons, and does not conflate symbol-use with the potential properties of an underlying symbolic framework. This view also implies that mechanisms that guarantee systematicity may be neither necessary nor sufficient for symbolic behaviour. Rather, systematicity may be a graded competency afforded by environmental and educational factors~\citep{mcclelland2010letting,hill2019environmental,mcclelland2020placing,nam2021underlies}. \\

Second, symbols are often defined as discrete, localist representations~\citep[e.g.]{garcez2020neurosymbolic}. That view fits with computers' binary representations, and thus allows leveraging computer algorithms in AI. Humans can interpret non-discrete entities as symbolic---for example, continuous variations in intonation, volume, and pitch of a single word can convey vastly different emotions, sarcasm, and so on. The important distinction is that a particular entity need not be discrete to be a member of a discrete set; the representation of a word can be continuous but treated as a discrete symbol, just as a continuous vector can be a member of a discrete set of basis vectors. If we are to appreciate symbolic thought in its most potent, general form, then we need to embrace the idea that symbols come in many shapes and forms. According to the definition presented here, any substrate can be symbolic to a fluent symbol user. The triadic relation of the substrate, the denotation, and the interpreter are all that is needed, not any particular properties of the substrate.}}

\section{Conclusion}

Classical perspectives on symbols in AI have mostly overlooked the fact that symbols are fundamentally subjective---they depend on an interpreter (or some interpreters) to create a convention of meaning. Thus, human-like symbolic fluency is not guaranteed simply because a system is equipped with classical “symbolic” machinery. Instead, symbolic fluency should be evaluated through behaviours, whether these behaviours involve interactions with interlocutors or simply demonstrate improved internal reasoning.This can be measured by inspecting a set of graded traits, such as receptiveness to new convention, the ability to construct new conventions, and demonstrated understanding of the meaning behind syntactic maneuvers. Because optimizing directly for behaviour is increasingly feasible, we argue that the key to developing machines with human-like symbolic fluency is to optimize learning-based systems for these symbolic behaviours directly by placing artificial agents in situations that require their active use. Human socio-cultural situations are perhaps best suited to fulfill this requirement, as they demand the complex coordination of perspectives to agree on arbitrarily-imposed meaning. They can also be collected at scale in conjunction with human feedback, and hence allow the use of powerful contemporary AI tools that mould behaviour.

\bibliographystyle{unsrtnat}
\nobibliography*
\bibliography{main}

\section*{Acknowledgements}
We thank Felix Hill, Matt Botvinick, Jay McClelland, Christos Kaplanis, Matt Overlan, and Ivana Kaji\'{c} for comments on the paper, and many others at DeepMind for the conversations over the years that have helped shape our views.

\end{document}